\documentclass{article}
\usepackage{ijcai22}

\usepackage{times}
\usepackage{cite}

\usepackage{soul}
\usepackage{url,color}
\usepackage[hidelinks]{hyperref}
\usepackage[utf8]{inputenc}
\usepackage[small]{caption}
\usepackage{graphicx}
\usepackage{amsmath}
\usepackage{booktabs}
\title{The BLue Amazon Brain (BLAB): \\  A Modular Architecture of Services about the Brazilian Maritime Territory}

\author{
Paulo Pirozelli$^{1,7}$\footnote{Contact and main author. All other authors contributed to specific
parts of the paper and are enumerated in alphabetical order with respect to first name, except the last five authors,
who oversaw the work and secured funding (and are also presented in alphabetical order).} \and
Ais B. R. Castro$^{2,7}$ \and
Ana Luiza C. de Oliveira$^{3,7}$ \and
André S. Oliveira$^{4,7}$ \and
Flávio N. Cação$^{4,7}$ \and
Igor C. Silveira$^{5,7}$ \and
Jo\~ao G. M. Campos$^{4,7}$ \and
Laura C. Motheo$^{2,7}$ \and
Leticia F. Figueiredo$^{3,7}$ \and
Lucas F. A. O. Pellicer$^{4,7}$ \and
Marcelo A. Jos\'e$^{1,7}$ \and 
Marcos M. Jos\'e$^{4,7}$ \and
Pedro de M. Ligabue$^{4,7}$ \and
Ricardo S. Grava$^{4,7}$ \and 
Rodrigo M. Tavares$^{4,7}$ \and
Vin\'icius B. Matos$^{7}$ \and 
Yan V. Sym$^{4,7}$ \and \\
Anna H. R. Costa$^{4,7}$ \and
Anarosa A. F. Brand\~ao$^{4,7}$ \and \\
Denis D. Mau\'a$^{5,7}$ \and
Fabio G. Cozman$^{4,7}$ \And
Sarajane M. Peres$^{6,7}$ \\
\affiliations
$^1$Instituto de Estudos Avan\c{c}ados, $^2$Curso de Ciências Moleculares, $^3$Instituto Oceanogr\'afico, $^4$Escola Polit\'ecnica, $^5$Instituto de Matem\'atica e Estat{\'\i}stica, $^6$Escola de Artes, Ci\^encias e Humanidades, \\ $^7$Center for Artificial Intelligence (C4AI) \\  \mbox{ } \\
Universidade de S\~ao Paulo, S\~ao Paulo - Brazil
}

\begin{document}

\maketitle

\begin{abstract}
We describe the first steps in the development of an artificial agent 
focused on the Brazilian maritime territory,
a large region within the South Atlantic also known as the {\em Blue Amazon}. 
The ``BLue Amazon Brain'' (BLAB) integrates a number of services aimed at disseminating information about this 
region and its importance, functioning as a tool for environmental awareness. The main service provided by BLAB is a conversational facility that deals with complex questions about the Blue Amazon, called BLAB-Chat; its central component is a controller that manages several task-oriented natural language processing modules (e.g., question answering and summarizer systems). These modules have access to an internal data lake as well as to third-party databases. A news reporter (BLAB-Reporter) and a purposely-developed wiki (BLAB-Wiki) are also part of the BLAB service architecture. 
In this paper, we describe our current version of BLAB's architecture (interface, backend, web services, NLP modules, and resources) and comment on the challenges we have faced so far, such as the
lack of training data and the scattered state of domain information.
Solving these issues presents a considerable challenge in the development of artificial intelligence for technical domains.

\end{abstract}

\section{Introduction}

There is a vast and ever-growing body of information about the oceans; 
clearly, society can benefit from artificial intelligence (AI) systems that provide 
services based on such information. 
At the same time, the ocean offers the perfect domain in which to test
these systems, given the wealth of background knowledge one must handle when 
processing ocean data. In this paper, we describe our preliminary efforts
towards an ``artificial brain'' focused on this domain. Admittedly, the word ``brain'' is a rather ambitious one, and the reader 
will notice that our current results do not authorize us to use this word 
with impunity. In any case, our goals are indeed ambitious: 
we wish to build an architecture that encapsulates a number of complex and interconnected services concerning ocean knowledge, from question answering to time series analysis. 
Our goal with this is to foster awareness about oceanographic issues --- from biodiversity to food supply, from energy resources to climate forecasts. 

\begin{figure}
    \centering
    \includegraphics[width=8.3cm,trim=0 0 0 46,clip]{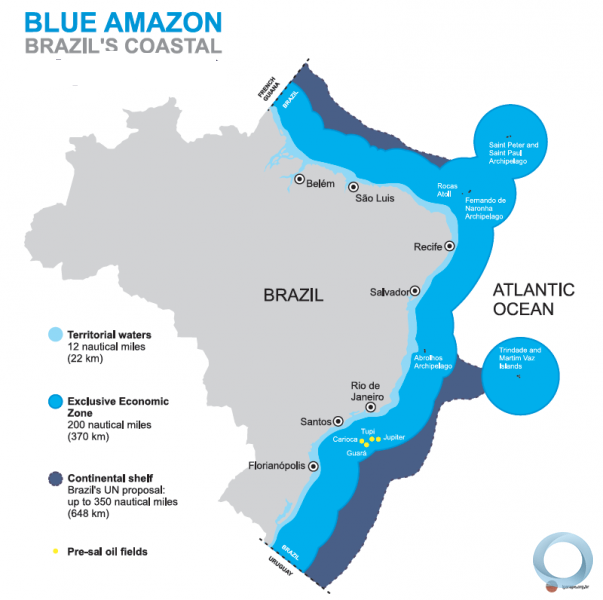}
    \caption{The Blue Amazon (extracted from \cite{foreignaffairs2015}).}
    \label{fig:blue_amazon}
\end{figure}

It would not be wise to expect that a computational brain could handle questions over 
the whole ocean; we thus focus on a specific part of it, 
namely the Brazilian maritime territory. 
This is a vast region of the South Atlantic, covering a few million
square kilometers, as indicated in Figure~\ref{fig:blue_amazon}.
Approximately as big as the Amazon rainforest, this region is often referred to as the \emph{Blue Amazon}.
Figure \ref{fig:blue_amazon} depicts the 
Brazilian territorial waters (up to 12 nautical miles from land),
the Brazilian exclusive economic zone (up to 200 nautical miles)
over which Brazil has sea and air sovereignty (the official Blue Amazon), and additional
territory of the continental shelf under discussion at the United Nations \cite{BrazilianNavy}. 
The Blue Amazon carries 95\% of Brazil's international trade and holds almost all of the country's (quite large) oil (95\%) and gas (80\%) reserves \cite{anp2021}. The region is also a vital source of food supply and a key player in climate regulation \cite{abreu2015}.

Yet the Blue Amazon is not well-known to the wider public or even to those living in the coastal region of Brazil. Information about it is dispersed in academic volumes and government reports, or in obscure databases. 
As such, the Brazilian maritime territory offers a perfect
domain for research on a variety of artificial intelligence themes: 
instead of relying on
curated and worn-out information, say  in Wikipedia, a Blue Amazon brain
must go through a maze of scattered information to come up with 
useful information to users. Obviously, this state of affairs makes our
goals even more challenging and requires us to move in small steps.

In this paper, we report the   steps we have already taken towards
our BLue Amazon Brain (a system we refer to as BLAB). 
In essence, we envision an architecture that encompasses a number of complex and interwoven services. The most important of these is a conversational agent that can take user requests,
from naive questions to highly technical issues,
and return relevant and accurate results. 
A key feature of this agent is responding to questions in a way that goes beyond
mere reproduction of stored templates. 
Such a question answering ability can be used to satisfy particular user
needs but also, more broadly, to educate the public about the domain.\footnote{One
might note the CliMate conversational agent at 
\url{https://davidsuzuki.org/climate-conversation-coach/}; another 
example is the SaveEcoBot that carries information about air quality at
\url{https://www.saveecobot.com/us}; in the Portuguese language, one can
find for instance the agent AGATA on water and energy waste  \cite{gomes2020agata}
and PipaBot on environmental pollutants  \cite{de2019pipabot}.}
Other useful services of BLAB are an automated news generator and a Blue Amazon wiki; additional
services are planned for the future. 
We also envision systems that can provide services 
such as predictions based on metocean data. 
Hopefully, BLAB can help solidify a conviction 
about the importance of the ocean to human life.

The main contributions of this paper are: a formal description of BLAB's architecture; an assessment of the natural language processing (NLP) modules we have developed so far, their main
limitations and our planned future developments; and a description of the resources we have created in connection with the ocean and in particular with the Blue Amazon.

The paper is organized as follows: Section \ref{sec:infrastructure} offers an overview of BLAB's architecture. Section \ref{sec:modules} goes over the specific NLP modules and
Section \ref{sec:resources} describes the resources we have developed so far. 
Section \ref{sec:final} concludes the paper with a summary of the challenges we have learned so far and the next steps we plan to give.

\begin{figure*}[ht]
\centering
  \includegraphics[width=17cm]{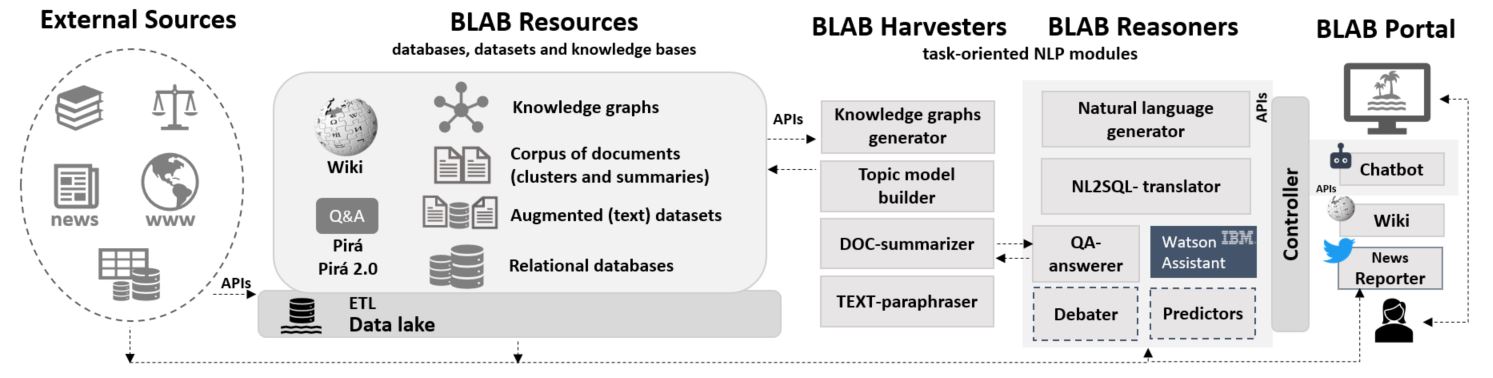}
  \caption{BLAB's overall architecture.}
  \label{fig:architeture}
\end{figure*}

This project is hosted by the Center for Artificial Intelligence (C4AI),\footnote{Information at \url{http://c4ai.inova.usp.br/}.}
a large research center, headquartered at the Universidade de S\~ao Paulo, that congregates 
researchers and students from a wide variety of fields. A broader goal of  this project
is to investigate how AI can benefit simultaneously
from data and knowledge-driven approaches. 

\section{An Overview of BLAB} \label{sec:infrastructure}


Figure \ref{fig:architeture} presents an overview of BLAB's architecture.
BLAB has three main components, namely a portal, a set of 
reasoners, and a set of resources. 
Originally, the portal was developed solely as a way of testing and validating 
the conversational agent. The portal was later expanded to encompass a number of other services related to the Blue Amazon, serving as a tool for disseminating knowledge on this domain to a wide audience.

Currently, the portal offers preliminary versions of three services: a conversational agent (BLAB-Chat), a wiki (BLAB-Wiki), and a news reporter (BLAB-Reporter). The conversational agent is managed by a central controller that calls several NLP modules 
in order to process requests. These modules can be roughly divided in two categories: \textit{reasoners}, which are directly 
connected to the dialogue process, working on the inputs received from the user (e.g., question answering system); and \textit{harvesters}, responsible for pre-processing and structuring data 
(e.g., multi-document summarizer). 
To process 
data, the reasoners have access to a data lake, containing data from multiple sources and of different types, structured and non-structured, such as textual documents and relational databases,
as well as to third-party services. The other two services, the reporter and the wiki, work independently. The BLAB-Reporter collects structured data from databases and generates reports based on them in natural language; it also summarizes news from the web. The BLAB-Wiki is our content-specific encyclopedia; it can be directly accessed by users through the website and is also mirrored in the data lake to serve as an expert information source for reasoners. 

From a technical standpoint, the operation of BLAB depends on a service-oriented infrastructure (back-end) and on a web interface (front-end) that accommodates multiple services and promotes user interaction. The next subsections present information about both BLAB layers, along with brief descriptions of the proposed web services.

\subsection{BLAB back-end}

BLAB's central controller is responsible for managing the 
communication between pairs of components. The controller is part of the back-end server, which we have implemented in Python using Django; thus, any SQL engine supported by Django (such as PostgreSQL and MySQL) can be used to manage data. The communication with clients is made through REST APIs and WebSockets created with Django REST Framework and Django Channels, respectively.

Currently, the controller’s role is to manage, mainly, services related to the dialog web application (chatbot).
However, its service-oriented architecture guarantees flexibility and maintainability, and simplifies the addition of new modules to the system. 
Our team plans, for example, to implement a framework for interactive maps with geo-referencing; the idea is to integrate this service with other BLAB reasoners, such as a tide predictor and a vessel tracker. 

Besides the 
controller, the back-end architecture contains a data lake for loading and storing files. The management is performed by a tool that updates files and folders from   Google Drive folders. The data lake stores data from various sources in their original format, from which system components may request specific files (extract-load-transform process). Ideally, when more data becomes available, data conversion routines will allow the direct access of processed data by the component systems (extract-transform-load process).

\subsection{BLAB front-end}

\begin{figure*}[ht]
\centering 
  \includegraphics[width=0.8\textwidth]{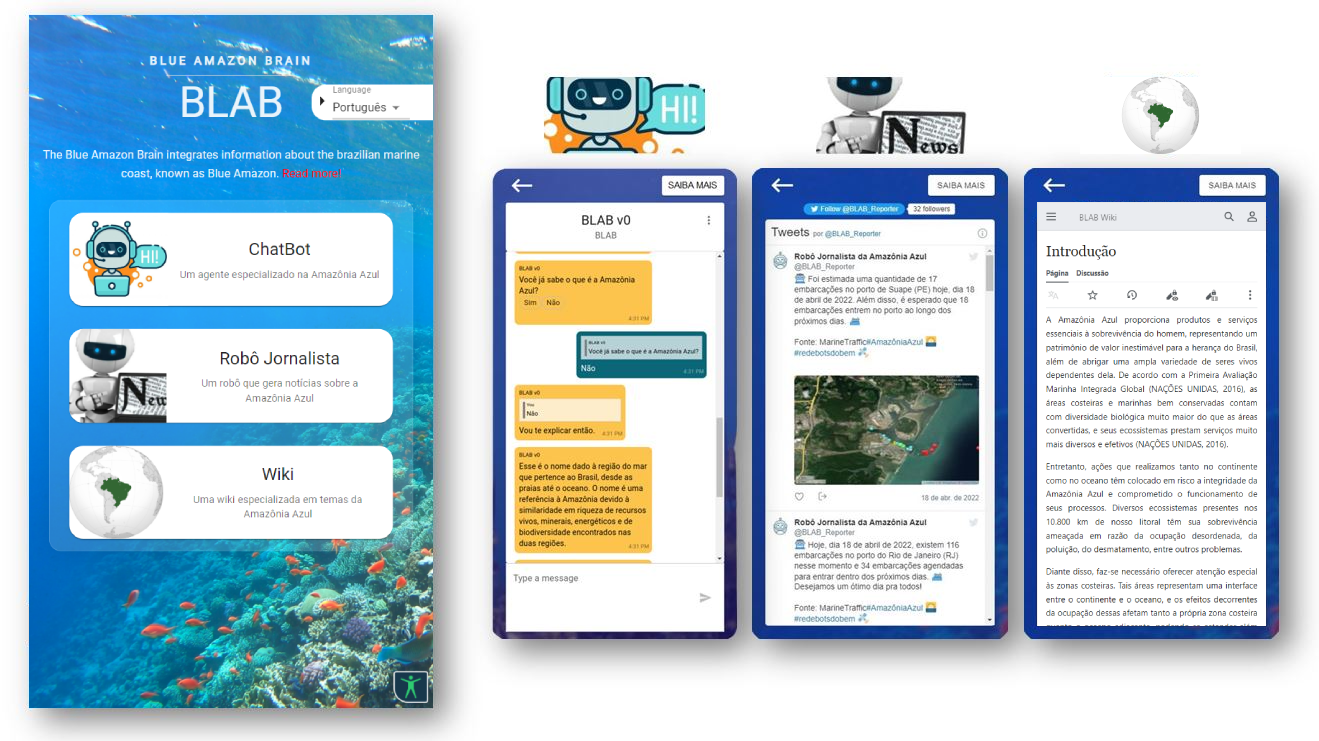}  \caption{On the left: emphasis on BLAB Portal. Next images, in sequence: BLAB-Chat, BLAB-Reporter and BLAB-Wiki.}
  \label{fig:website-chatbot}
\end{figure*}

Public access to BLAB's features is done through a web interface to provide a portal of intelligent services for the dissemination of information about the Blue Amazon. It was developed to allow the integration of new services on demand. Currently, it hosts preliminary versions of three services: a chatbot, a robot reporter, and a wiki, all specialized in Blue Amazon. Pieces of an interface screenshot can be seen in Figure~\ref{fig:website-chatbot}. These services are currently available only for internal testing. 


The interface conception considered four fundamental requirements to provide good user experience while interacting with BLAB. These requirements are: (i) \textit{internationalization}; (ii) \textit{modularity}; (iii) \textit{maintainability}; and (iv) \textit{accessibility}. Having an interface compliant with them, we guarantee: (i) its use by speakers of different languages; (ii) and (iii) easy integration of new services and maintenance of the existing ones due to the adoption of a code style organization and associated documentation; and (iv) universal access to BLAB services, independently of   user limitations.

Technically, the web interface has been developed with the React library, along with the MUI component library. These libraries simplify the development process, as they allow to generate dynamic single-page applications through the exclusive use of Javascript and pre-supplied templates. To facilitate internationalization, the interface uses the i18next-react\footnote{https://react.i18next.com/} framework, which simplifies the translation of textual elements of the website into different languages, by simply providing their translation in JSON files. 
To comply with accessibility standards, the four basic principles established by by WCAG have been adopted.\footnote{\url{https://www.accessibility.works/blog/2022-ada-wcag-website-accessibility-standards-requirements/}} Such principles state that a website must be perceptible, operable, understandable, and robust. \textit{Perceptibility} means that all users must perceive web interface elements with their senses, which involves creating presentable content in ways that facilitate perception, such as providing alternative texts for non-text content. To guarantee \textit{operability}, the interface cannot require interactions that a user cannot perform, so one requirement is to allow the use of all possible functionalities from a keyboard. \textit{Understandability} requires all information to be understandable; i.e., text must be readable and clear, and content must be presented predictably. Finally, \textit{robustness} requires that content must be interpretable by a variety of user agents, including assistive technologies such as 
screen-reading tools. Our efforts have complied with these principles.

\subsection{BLAB web services} \label{sec:services}

Three web services are currently in operation in the BLAB web interface. 
Details on each 
of them are provided below. 

\paragraph{BLAB-Reporter:} The main goal of BLAB is to raise public awareness of ocean-related issues, thus affecting how the population perceives and interacts with the environment. One way to do that is by reporting news about the oceanic domain \cite{UNESCO:2021}. Unfortunately, most data-stream sources are available only in numerical or machine-readable formats, preventing wider audiences to understand them. 
To overcome that, we designed the 
BLAB-Reporter,\footnote{\url{https://twitter.com/BLAB_Reporter}} an application that
collects data related to the Blue Amazon and publishes it on Twitter in natural language, 
and we integrated it with BLAB Portal. The use of Twitter 
aims to bring this marine-related content to broader audiences.
 The
BLAB-Reporter produces tweets with news through a traditional natural language generating pipeline (details are available in Section~\ref{sec:modules}) using accurate real-world data. For future development, we plan to add other discourse intentions to the automated journalist, such as fishing activity watching, oil extraction, spills monitoring, and endangered marine species tracking.

\paragraph{BLAB-Wiki:} BLAB provides a service to access our Blue Amazon wiki (still in a preliminary version). The content is stored in the internal data lake and is mirrored through a web view. The purpose of the wiki in BLAB's architecture is twofold: disseminating knowledge through BLAB Portal, and serving as  an important resource for providing content to BLAB reasoners (Subsection \ref{sec:blab-wiki}).

\paragraph{BLAB-Chat:} BLAB's portal provides a generic chat service framework that supports many types of messages (e.g., text, voice recording, file attachment), as well as basic chat features such as timestamps and message quoting; this is also in a preliminary version.  
The chatbot, whose chatting features rely on IBM's Watson Assistant, 
is responsible for collecting answers from the QA modules (discussed later) and shipping 
them to the portal.\footnote{%
Note that IBM Watson is a computer system developed by IBM Research with the aim of competing on the American TV show \textit{Jeopardy} against human contestants in real time \cite{watson10};
it led to the development of IBM Watson Assistant (\url{https://cloud.ibm.com/catalog/services/watson-assistant}), a service that allows the creation of conversational agents that can be embedded in any application.}

\section{Task-oriented NLP modules}\label{sec:modules}
The backbone of BLAB is a conversational agent, the BLAB-Chat. Its goal is to directly engage with users, raising their interest and understanding about the Blue Amazon. As mentioned in Section \ref{sec:infrastructure}, BLAB also includes a reporter functionality that publishes tweets. Both the conversational agent and the reporter need to understand and generate natural language to properly interact with users. 
Thus, to endow BLAB with natural language processing capabilities, we have  developed a set of NLP modules,\footnote{Available at \url{https://github.com/C4AI}.} which can be broadly divided into \textit{reasoners} and \textit{harvesters}.
Reasoners work with the controller directly, providing the required information to the chatbot and the reporter. Currently, this group of modules comprise a natural language generator; a natural language to structured query language (NL2SQL) translator; an open question answering (QA) system; and a third-party service to implement the dialog flow, Watson Assistant from IBM. Harvesters create and structure content about the Blue Amazon domain; this may be done in real time, when called by reasoners, or by directly feeding the data lake. This set of modules is currently formed by a knowledge graph generator; an unsupervised topic model builder; a multi-document summarization module; and a paraphraser. 

\subsection{BLAB Reasoners}

In this subsection, we present a brief summary of the existing reasoners.

\paragraph{Natural language generator:} Most information about the Blue Amazon   can be found in structured databases or in unstructured data repositories. These data 
may be used to feed systems which automatically generate reports \cite{Teixeira:2020}. BLAB-Reporter is a data-to-text natural language generation model that publishes real-time recurrent content in a human-readable fashion \cite{Ferreira:2019}, by acquiring data from various web sources (information concerning marine and coastal weather, tide charts, marine vessel traffic, and news regarding the  Blue Amazon).  
To prepare content for publication, the system follows a pipeline with six steps: content selection; discourse ordering; text structuring; lexicalization; referring expression generation; and textual realization. Through these steps, the original non-linguistic data passes through a series of intermediate representations, selecting relevant information, organizing it in logical order, determining the proper lexical choices, until finally rendering the text in its final form \cite{Campos:2020}. These steps follow a template-based format and draw their choices from a list of possibilities built from annotated examples. The use of a rule-based approach ensures high-fidelity for numerical report texts; however, this is done at the expense of textual diversity. To overcome that limitation, we are working on data-driven approaches to the lexicalization and referring expression generation steps of the pipeline. Our aim with that is to provide greater variability and fluency to the text, thus enhancing audience engagement.

\paragraph{NL2SQL translator:} Structured data can be naturally organized in relational databases. For this type of data, SQL offers an efficient approach for information retrieval. A conversational agent that is able to translate query-type user interactions into SQL statements can recover faster and more robust answers.
To implement this sort of interaction, BLAB includes a module that translates natural language into SQL. Our approach is based on the RAT-SQL+GAP~\cite{RAT-SQL+GAP} architecture, which joins two components: the Relation-Aware Transformer SQL (RAT-SQL)~\cite{RAT-SQL}, which encodes the relations between a natural language request and a domain-oriented database schema; and the Generation-Augmented Pre-training (GAP), a BART model \cite{lewis2019bart} pretrained for SQL and text-related tasks, and then specifically fine-tuned for NL2SQL using RAT-SQL. 

When developing the NL2SQL translator for BLAB, we faced the challenge of working with Portuguese, for which no model was yet available. Thus, our work focused mainly on the adaptation of the RAT-SQL+GAP architecture to Portuguese. This required the implementation of UTF-8 encoding in the RAT-SQL component, in order to support characters from languages other than English, as well as multilingual models such as mBART-50~\cite{mBART50} and mT5~\cite{mT5}. We named this language-agnostic model \textit{mRAT-SQL}. With these adaptations, we were able to fine-tune our model to Portuguese \cite{mRAT-SQL}. In addition, we applied a data augmentation strategy based on back-translations of the RAT-SQL+GAP training dataset into several languages. Finally, in order to promote an integration with a QA system, we trained a text classifier using BERTimbau~\cite{souza2020bertimbau}, a pre-trained BERT \cite{devlin2018bert} model for Portuguese. This classifier produced good results in distinguishing types of questions~\cite{IntegratinQA-NL2SQL}. We are currently working to fully attach it to BLAB's architecture, so it can differentiate questions by type and send them to the appropriate QA modules.

\paragraph{QA system:} \label{sec:deepage}
The first module developed for BLAB was an end-to-end question answering module, called DEEPAGÉ \cite{deepage}. Given a question in Portuguese, this module is able to traverse a corpus of documents and gather relevant information, returning 
an appropriate answer in the same language. To choose the best architecture for the QA system, we compared several combinations of BM25 \cite{Robertson2009}, a sparse retriever, and PTT5 \cite{carmo2020ptt5}, a T5-based \cite{raffel2019exploring} reader pre-trained on Portuguese textual databases: a retriever-only system (BM525); a reader-only system (PTT5); and a dual system (BM25 + PTT5). We also experimented pre-trained and fine-tuned versions of PTT5. In the dual system, the retriever component is responsible for inspecting the corpus and finding the $k$ most relevant documents to answer a user's query, and the reader component is responsible for turning these $k$ documents into a final answer. The QA fine-tuning of PTT5 was done with questions from PAQ, a massive open domain QA dataset with 65M question-answer pairs (QA-pairs) \cite{lewis-etal-2021-paq}, filtered for questions related to Brazil's environment based on a set of hand-crafted regular expressions. The process resulted in a QA dataset with 14K instances, which were 
translated into Portuguese with the Google Translation API. As for the reader, we assembled our own corpus of documents. For that, we downloaded and filtered all articles in the category ``Environment of Brazil'' from the Wikipedia in Portuguese (17K articles), and scraped Brazil's three biggest newspapers within a window of three and a half years, using relevant keywords to find articles on environment (29K articles). As expected, the best result was achieved by the dual system, whose reader component has been fine-tuned on our dataset.
This result shows the importance of having a corpus and a QA dataset focused on the Blue Amazon, as well as the gains of fine-tuning PTT5 for questions within this domain. After developing DEEPAGÉ, we fine-tuned it on our Blue Amazon-oriented corpus and QA dataset, Pirá 2.0 (Section \ref{sec:resources}). 

\subsection{BLAB Harvesters}

In this subsection, we present a brief summary of existing harversters.

\paragraph{Knowledge graph generator:} \label{sec:graph}
Knowledge graphs represent entities and the relationships between them. Their usefulness comes from the fact that they provide a defined and direct way of encoding knowledge as a network, which can then be traversed and expanded in a straightforward manner. This makes knowledge graphs useful for enriching QA systems, in which questions provide a starting point and directions, and answers must be found elsewhere within the knowledge base \cite{kg_cloud_education_2021}. In the context of BLAB, our intention is to create one or more consistent knowledge graphs from the various independent documents stored in the data lake. The resulting graphs may be used as a first step in formalizing knowledge, such as in an ontology, or as a more transparent way to inspect symbolic representations and reasoning steps. 

We have developed a knowledge graph generator that can be viewed as an extension of AutoKG \cite{autokg}. This method comprises four steps. First, sets of triples in the format \textit{$\langle$subject, relation, object$\rangle$} are extracted from the documents in the corpus through the OpenIE annotator \cite{openie}. Second, a graph is built for each individual document by going through the triples and identifying entities that can be understood as synonyms (this identification is done by comparing contextual embeddings generated with BERT for each of the entities in the triples). Third, graphs are narrowed down by reducing the group of synonyms to the most recurring entity. Fourth, entities in separate graphs are linked together through the comparison of word embeddings. In the latter step, links between entities are not treated as synonyms, but as ``bridges'' between different contexts (i.e., documents). 

\paragraph{Unsupervised topic model builder:} \label{sec:unsupervised}

The documents in our knowledge bases go over a wide variety of topics. Treating them as a single block is an ineffective, and perhaps unfeasible, strategy. An alternative to that is using some sort of clustering procedure to model topics of interest, which can offer clues on which documents to look when a specific subject appears on the chatbot agenda.

In order to model such topics, we employed a co-clustering method. Besides discovering additional information in the form of the co-cluster structures, co-clustering algorithms can frequently find better clusters than standard one-sided clustering techniques \cite{dhillon2003information}. Non-negative Matrix Tri-Factorization (NMTF) \cite{yoo2010orthogonal} was chosen as our co-clustering algorithm, due to its simplicity, performance and frequent use in textual analysis \cite{salah2018word}. NMTF was carried out on the set of abstracts of Pirá dataset. After co-clustering, the factor matrices were post-processed   to achieve additional semantic explanations to the co-cluster structure by interpreting row and column clusters, and their association, as document and word clusters respectively (cf. suggested in \cite{de2020ovnmtf}). This procedure resulted in each abstract being assigned to a topic which can be summarized by a set of words, allowing us to interpret the underlying material that composes the supporting texts of Pirá.

Associating a text with a topic considerably speeds up the retrieval of documents and specializes the knowledge derived from them. For a real-time operation, such as a conversation, this may represent a considerable time reduction and the construction of more precise answers. The effects of using topic modeling on the accuracy and efficiency of reasoners are part of the next testing steps in BLAB's development. Still, there is considerable room for future exploration in our approach. For instance, the co-clusters structures can be used to analyze overlaps between clusters to show relations between topics; or to measure how representative the associations between words and documents within a topic are, 
opening an opportunity to suggest the basis for argumentation for a debater module.

\paragraph{Multi-document summarizer:} Even when documents are organized into smaller groups, scanning full texts may still represent a considerable problem. Larger documents, such as books, may contain hundreds of thousands of words. 
Then, not only unnecessary time may be spent in irrelevant parts, but QA models may simply drown with that amount of information. Thus, in order to dispose information about the Blue Amazon in a more concise format, we developed a multi-document abstractive summarization model for Portuguese called \textit{PLSUM} \cite{plsumeniac}. 

The multi-document abstractive summarization model receives a summary title as query---which could be, for example, a question made by a user. Next, it accesses a textual corpus, retrieving multiple sentences related to the title. Finally, it uses the sentences to produce an authorial, non-extractive, summary about the title. Another way of seeing a multi-document abstractive summarization is as a system having two major stages: an extractive model that extracts the similarity of the document sentences to the title and outputs the best $L$ sentences in order of relevance; these $L$ sentences are then concatenated with the title and passed through an abstractive model, which generates a summary with a maximum size of $n$ tokens.

To find the best combination of extractive and abstractive components for our summarization task, we tested several candidates and compared them according to standard metrics. For the extractive stage, which infers the relevance of sentences, we tested some variations of the TF-IDF \cite{ramos2003using}. On the abstractive stage, we compared fine-tunings of two encoder-decoder transformers: PTT5 and Longformer \cite{beltagy2020longformer}. To fine-tune the abstractive stage models, we created a new dataset in Portuguese, the BrWac2Wiki \cite{plsumeniac}. Each sample of the
dataset associates a title and set of documents extracted from the internet (input) with a Wikipedia lead (target summary). As a continuation of this work, we intend to specialize our model in generating summaries about environmental topics, so that it can produce more accurate results in the Blue Amazon domain. We are also working on the improvement of the extractive stage, as to reduce inaccuracy and redundancy.

\paragraph{Paraphraser:} \label{sec:paraphraser}

As some of our experiments in the NL2SQL translator have demonstrated, data augmentation can be a valuable tool in NLP applications. By enlarging datasets, it is possible to reduce overfitting, thus allowing the training of bigger models \cite{li2022data}. Data augmentation is particularly useful for low-resource languages and closed domains \cite{Fadaee2017}, two scenarios that apply to BLAB. One particularly interesting form of data augmentation in NLP is \textit{paraphrase generation}. Paraphrasing methods can be broadly divided into three groups: lexical approaches \cite{miller95,ganitkevitch-etal-2013-ppdb}; back-translation approaches \cite{Mallinson2017}; and mixed methods \cite{Hu2019}. In all these cases, there seems to be a trade-off between meaning preservation and diversity. 

As an attempt to overcome such a difficulty, we have developed our own paraphraser, PTT5-Paraphrase \cite{pellicer2022ptt5}. PTT5-Paraphrase is a PTT5 model fine-tuned on the Portuguese part of TaPaCo \cite{scherrer2020tapaco}, a large corpus collection of paraphrases in several languages. We compared PTT5-Paraphrase to other approaches according to three metrics: METEOR, BLEU (without brevity penalty), and cosine dissimilarity of sentence embeddings (using Sentence-BERT for encoding sentences \cite{reimers2019sentence}). On the one hand, PTT5-Paraphrase scores much better than rule-based approaches, such as WordNet \cite{miller95} and PPDB \cite{ganitkevitch-etal-2013-ppdb}, as regards diversity (lower METEOR/BLUE). On the other hand, PTT5-Paraphraser is considerably better in preserving meaning (lower cosine dissimilarity) than other neural approaches, such as ParaNet \cite{Mallinson2017} and Parabank \cite{Hu2019}. Overall, it achieves a good compromise between diversity and semantic fidelity. PTT5-Paraphraser was then validated on the manual paraphrases from Pirá (Subsection \ref{sec:pira_1}). No statistical difference in meaning preservation and clarity has been detected, although automatically-generated paraphrases were considerably less creative. Currently, we are working to integrate PTT5-Paraphrase into BLAB's architecture; the paraphraser will be used to augment existing datasets on a regular basis, thus improving the training and fine-tuning of reasoners.

\section{Resources}\label{sec:resources}
In this section, we describe the main resources produced for BLAB: two versions of a QA dataset (Pirá and Pirá 2.0), a domain-specific wiki, and a large corpus of documents on the Brazilian maritime territory. Currently, we are also working to populate the data lake with knowledge graphs, text summaries, topic-oriented clusters of texts, and augmented datasets, resources that will be produced by the harverster modules described in Section \ref{sec:modules}. 

\subsection{Pirá and Pir\'a 2.0} Pirá is a bilingual (Portuguese-English) reading comprehension dataset about the ocean, the Brazilian coast, and climate change \cite{paschoal2021pira}. The dataset consists of 2,261 question/answer (QA) sets in both languages. To the best of our knowledge, Pirá is the first open-ended QA dataset with questions in Portuguese, and, perhaps more importantly, the first bilingual QA dataset that includes this language. QA sets were manually created based on two corpora: one with scientific abstracts related to the Brazilian coast and the other with excerpts from two United Nation reports about the ocean \cite{un2017world, un2021world}. Next, the QA sets were evaluated in a peer-review process according to several aspects, such as meaningfulness, difficulty, and question type. As part of the assessment, volunteers answered the original questions too, thus providing a natural human baseline for the dataset. Volunteers also produced paraphrases of the original questions, which were later used to validate our paraphraser (Subsection \ref{sec:paraphraser}).

\label{sec:pira_1}

In   subsequent work, we defined five benchmarks for Pirá: machine reading comprehension, information retrieval, open question answering, answer triggering, and multiple choice question answering. For each task, we obtained a number of baselines, including human (when available), random, and NLP models' performance, measuring them against standard metrics. As part of our effort, we also produced a curated version of the original dataset, Pirá 2.0, obtained by a thorough revision for grammar issues, repeated questions, and other shortcomings. 

Furthermore, the dataset was extended in a number of new directions required by baselines, such as multiple choice candidates, classification labels, and automatic paraphrases. Multiple choice candidates were created by extending the original QA sets from Pirá. Each multiple choice set has five alternatives: the correct answer to a question and the answers to four other questions in the dataset. The distractors were selected based on their similarity with the target text, as to make them as plausible as possible. To create the answer triggering dataset, we used the results obtained in the evaluation phase of Pirá. In this stage, participants had to evaluate the meaningfulness of a question in a Likert scale (1--5). Assuming that a meaningless question is one for which there is no answer, we converted the evaluations to 0 (not answerable) or 1 (answerable) based on a threshold. Finally, automatic paraphrases were generated with our paraphraser (Subsection \ref{sec:paraphraser}), which had been validated on the original dataset.

Each of the benchmarks represents a challenge towards developing a program that can interact with users by answering their questions. Information Retrieval deals with fetching relevant texts. Machine reading comprehension is responsible for generating an answer based on a text which contains the desired information. Open question answering extends this problem to questions with non-extractive answers. Answer triggering is responsible for deciding whether a question should be answered or not. Finally, multiple choice depends on finding the correct answer among a set of candidates; a task that, in our case, simulates the management of multiple question answering systems, in which a single answer has to be selected.

Pirá 2.0 is the main resource used by BLAB's NLP modules. In the reasoner side, it has been used in the fine-tuning of DEEPAGÉ, our main QA system (Subsection \ref{sec:deepage}). The corpus of supporting texts was used both in the training of the knowledge graph generator (Subsection \ref{sec:graph}) and the unsupervised topic model builder (Subsection \ref{sec:unsupervised}).


\subsection{BLAB-Wiki} \label{sec:blab-wiki}
One of the main issues we have faced in the development of BLAB modules was the lack of comprehensive texts on basic topics related to the Blue Amazon. Although   specialized material can be found in books and articles,   accessible texts on the subject are rarely available. For this reason, we decided to develop our own small wiki, named BLAB-Wiki, written by experts in the field.
The entries for this encyclopedia present basic topics concerning the Blue Amazon 
for lay readers based on the scientific literature. These texts serve as summaries of scientific topics, as an entryway to the general public, and as a friendly link to the technical literature. BLAB-Wiki's content is organized into four main axes: socio-environmental, biodiversity, physicochemical, and legislation and governance. Each axis is composed of entries discussing different aspects of the Blue Amazon, such as marine pollution, effects of climate change on the ocean, and deep sea ecology. BLAB-Wiki will serve as a reliable resource for the various modules of BLAB, such as the QA and summarizer systems. Its content is   revised by scientific consultants, and will soon be available in two formats: both as an ebook and as the wiki service that can be directly accessed through the BLAB's web portal (Subsection \ref{sec:services}).

\subsection{Corpus of documents}
In order to allow text queries, our data lake gathers a large number of documents on topics related to the Brazilian coast. Among these, we may cite: intergovernmental reports on the ocean, such as those by the UN and UNESCO; reports from the Brazilian government, in particular from the Ministry of the Environment; academic books, such as ``Noções de Oceanografia'', by the Oceanographic Institute of USP, and ``Brazil and the Sea in the 21st Century'', by the Center of Excellence for the Brazilian Sea (Cembra); scientific articles, theses, and dissertations from various sources; and legal documents with the main frameworks involving the Brazilian maritime territory, such as the Decree 5,300 from 2004, which established the National Coastal Management Plan.
Due to copyright issues, this corpus is not directly accessible to users, but only to BLAB's modules through the data lake. As to avoid any type of plagiarism or misappropriation, we are developing a tool to reference the original sources from where answers were primarily obtained.

\section{Conclusion: Lessons Learned, Future Work} \label{sec:final}

We have described in this paper our efforts in building an architecture of complex and interconnected services about the Brazilian maritime territory, the BLue Amazon Brain (BLAB). The  main BLAB service is a conversational agent, the BLAB-Chat, with advanced reasoning and user interaction capabilities. 
The core components of this architecture are
a controller that coordinates a data lake, an interface, and several NLP modules.
The latter include a natural language generator,
a QA system, a natural language to SQL converter, a knowledge graph generator, 
an unsupervised topic model builder, a multi-document summarizer, and a paraphraser.
BLAB also offers services of automated news (the BLAB-Reporter) and a thematic wiki (the BLAB-Wiki).
The whole system relies on a set of resources that include two QA datasets on the Blue Amazon, a large corpus of documents, and our wiki. 


A large system such as BLAB, and particularly the BLAB-Chat, exercises many aspects of 
current artificial intelligence. Although the required techniques, such as natural language processing and question answering, are getting better every day and are able to display 
surprising performance, not everything is perfect; thus, we would like to 
share a few lessons we have learned in the journey:



\begin{itemize}
\item 
Data-driven models, such as transformers and the like, are trained on massive amounts of data which endows them with some linguistic and common sense skills. Nonetheless, they still demand considerable efforts to learn how to handle specialized domains, such as the Blue Amazon: additional data must be collected,
corpora must be carefully curated, and optimization runs must be endlessly repeated, often
to frustrating results. 
Most data-driven models do not perform well when fed with  scattered knowledge
sources. Better techniques to incorporate specific information are sorely needed in AI.

\item There are clear limitations to NLP models when one deals with languages other than English: good trained models are harder to find and data is much more scarce. That led us to build our own resources (aimed at the Portuguese language)
when putting together BLAB. Even though large multilingual models can interpolate significant information across languages, 
it is important to have models and resources directly available in low-resource languages as well, particularly when one is concerned with a topic of universal interest such as the ocean.
\end{itemize}

In the near future we expect to expand the number of NLP-based
modules and resources in BLAB.
For instance, we plan to develop other types of reasoner modules, such as a debater and a sequential planner, which 
can provide the conversational agent with argumentation skills.
We are also working on adding prediction modules based on numeric data, such as 
tide and ocean current forecasters. These modules will generate information that will be useful for people who work, live or intend to carry out some social or economic activity in the Blue Amazon.
Besides expanding our current wiki, we hope to create 
a graph-based question answering dataset 
and an ontology of the Blue Amazon. As the set of resources related to the Blue Amazon grows, NLP modules will get better and easier to train. 
We hope that BLAB can become a useful tool for environmental education
and awareness, providing reliable and engaging content on the Blue Amazon to a wide range of users.

\section*{Acknowledgement}

The authors would like to thank Gabriel Okamoto Carlos, Caio Fabricio Deberaldini Netto, and Caio Noboru Asai for their previous contribution to this project.

The authors thank the Center for Artificial Intelligence (C4AI-USP) and the support from the São Paulo Research Foundation (FAPESP grant \#2019/07665-4) and from the IBM Corporation. A.\ L.\ C.\ de Oliveira, L.\ C.\ Motheo, L.\ F.\ F., and R.\ M. Tavares were supported by the Programa Institucional de Bolsas de Iniciação em Desenvolvimento Tecnológico e Inovação (PIBITI) of CNPq; R.\ S.\ Grava was supported by the Programa Institucional de Bolsas de Iniciação Cientifica (PIBIC) of CNPq; and A.\ B.\ R.\ Castro was supported by the Programa Unificado de Bolsas de Estudos para Apoio e Formação de Estudantes de Graduação (PUB) of the Universidade de São Paulo. This research was also partially supported by Itaú Unibanco S.A.; M.\ M.\ José, F.\ Nakasato and A.\ S.\ Oliveira have been supported by the Itaú Scholarship Program (PBI) of the Data Science Center (C2D) of the Escola Politécnica da Universidade de São Paulo. A.\ H.\ R.\ Costa and F.\ G.\ Cozman and D. D. Mau\'a thank the support of the National Council for Scientific and Technological Development of Brazil (CNPq grants \#310085/2020-9 and \#312180/2018-7 and 304012/2019-0, respectively).

\bibliographystyle{abbrv}
\bibliography{ijcai22}

\begin{thebibliography}{10}

\bibitem{abreu2015}
C.~T. Abreu.
\newblock Brazilian coastal and marine protected areas importance, current
  status and recommendations.
\newblock {\em Division for Ocean Affairs and the Law of the Sea, Office of
  Legal Affairs, The United Nations New York}, 2015.

\bibitem{openie}
M.~Banko, M.~J. Cafarella, S.~Soderland, M.~Broadhead, and O.~Etzioni.
\newblock Open information extraction from the web.
\newblock In {\em Proceedings of the 20th International Joint Conference on
  Artificial Intelligence}, IJCAI'07, page 2670–2676, San Francisco, CA, USA,
  2007. Morgan Kaufmann Publishers Inc.

\bibitem{beltagy2020longformer}
I.~Beltagy, M.~E. Peters, and A.~Cohan.
\newblock Longformer: The long-document transformer.
\newblock {\em arXiv preprint arXiv:2004.05150}, 2020.

\bibitem{BrazilianNavy}
M.~Brasileira.
\newblock Amazônia azul.
\newblock {\em Available at: www.marinha.mil.br/delareis/?q=amazoniazul},
  Accessed on 31 May 2022.

\bibitem{deepage}
F.~N. Ca{\c{c}}{\~a}o, M.~M. Jos{\'e}, A.~S. Oliveira, S.~Spindola, A.~H.~R.
  Costa, and F.~G. Cozman.
\newblock Deepag{\'e}: Answering questions in portuguese about the brazilian
  environment.
\newblock In A.~Britto and K.~Valdivia~Delgado, editors, {\em Intelligent
  Systems}, pages 419--433, Cham, 2021. Springer International Publishing.

\bibitem{Campos:2020}
J.~Campos, A.~Teixeira, T.~Ferreira, F.~Cozman, and A.~Pagano.
\newblock Towards {Fully} {Automated} {News} {Reporting} in {Brazilian}
  {Portuguese}.
\newblock In {\em Anais do {XVII} {Encontro} {Nacional} de {Inteligência}
  {Artificial} e {Computacional}}, pages 543--554, Porto Alegre, Brasil, Oct.
  2020. SBC.
\newblock ISSN: 0000-0000 Place: Evento Online.

\bibitem{carmo2020ptt5}
D.~Carmo, M.~Piau, I.~Campiotti, R.~Nogueira, and R.~Lotufo.
\newblock Ptt5: Pretraining and validating the t5 model on brazilian portuguese
  data.
\newblock {\em arXiv preprint arXiv:2008.09144}, 2020.

\bibitem{de2020ovnmtf}
W.~L. de~Freitas, S.~M. Peres, V.~Freire, and L.~F. Brunialti.
\newblock Ovnmtf algorithm: an overlapping non-negative matrix
  tri-factorization for coclustering.
\newblock In {\em 2020 International Joint Conference on Neural Networks
  (IJCNN)}, pages 1--8. IEEE, 2020.

\bibitem{de2019pipabot}
L.~S. de~Paula, T.~G. Gon{\c{c}}alves, T.~V. Fernandes, and G.~H. Travassos.
\newblock Pipabot: um canal de comunica{\c{c}}{\~a}o para o pipa ufrj.
\newblock In {\em Anais Estendidos do XXV Simp{\'o}sio Brasileiro de Sistemas
  Multim{\'\i}dia e Web}, pages 103--107. SBC, 2019.

\bibitem{devlin2018bert}
J.~Devlin, M.-W. Chang, K.~Lee, and K.~Toutanova.
\newblock Bert: Pre-training of deep bidirectional transformers for language
  understanding.
\newblock {\em arXiv preprint arXiv:1810.04805}, 2018.

\bibitem{dhillon2003information}
I.~S. Dhillon, S.~Mallela, and D.~S. Modha.
\newblock Information-theoretic co-clustering.
\newblock In {\em Proceedings of the ninth ACM SIGKDD international conference
  on Knowledge discovery and data mining}, pages 89--98, 2003.

\bibitem{anp2021}
A.~N. do~Petróleo~(ANP).
\newblock Encarte de consolidação da produção 2021: Boletim da produção
  de petróleo e gás natural.
\newblock {\em Available at:
  https://www.gov.br/anp/pt-br/centrais-de-conteudo/publicacoes/boletins-anp/boletins/arquivos-bmppgn/2021/12-2021-boletim.pdf},
  Accessed on 31 May 2022.

\bibitem{Fadaee2017}
M.~Fadaee, A.~Bisazza, and C.~Monz.
\newblock {D}ata {A}ugmentation for {L}ow-{R}esource {N}eural {M}achine
  {T}ranslation.
\newblock In {\em Proceedings of the 55th Annual Meeting of the Association for
  Computational Linguistics (Volume 2: Short Papers)}. Association for
  Computational Linguistics, 2017.

\bibitem{Ferreira:2019}
T.~C. Ferreira, C.~van~der Lee, E.~Van~Miltenburg, and E.~Krahmer.
\newblock Neural data-to-text generation: A comparison between pipeline and
  end-to-end architectures.
\newblock {\em arXiv preprint arXiv:1908.09022}, 2019.

\bibitem{watson10}
D.~A. Ferrucci, E.~W. Brown, J.~Chu{-}Carroll, J.~Fan, D.~Gondek, A.~Kalyanpur,
  A.~Lally, J.~W. Murdock, E.~Nyberg, J.~M. Prager, N.~Schlaefer, and C.~A.
  Welty.
\newblock Building {W}atson: An overview of the {DeepQA} project.
\newblock {\em {AI} Mag.}, 31(3):59--79, 2010.

\bibitem{ganitkevitch-etal-2013-ppdb}
J.~Ganitkevitch, B.~Van~Durme, and C.~Callison-Burch.
\newblock {PPDB}: The paraphrase database.
\newblock In {\em Proceedings of the 2013 Conference of the North {A}merican
  Chapter of the Association for Computational Linguistics: Human Language
  Technologies}, pages 758--764, Atlanta, Georgia, June 2013. Association for
  Computational Linguistics.

\bibitem{gomes2020agata}
B.~R. Gomes, A.~F.~L. Jacob~Jr, I.~d. J.~P. Pinto, and S.~Colcher.
\newblock {\'A}gata: um chatbot para difus{\~a}o de pr{\'a}ticas para
  educa{\c{c}}{\~a}o ambiental.
\newblock In {\em Anais Estendidos do XXVI Simp{\'o}sio Brasileiro de Sistemas
  Multim{\'\i}dia e Web}, pages 85--89. SBC, 2020.

\bibitem{Hu2019}
J.~E. Hu, R.~Rudinger, M.~Post, and B.~V. Durme.
\newblock {PARABANK}: Monolingual bitext generation and sentential paraphrasing
  via lexically-constrained neural machine translation.
\newblock In {\em The Thirty-Third {AAAI} Conference on Artificial
  Intelligence, {AAAI} 2019}, volume~33, pages 6521--6528. Association for the
  Advancement of Artificial Intelligence ({AAAI}), jul 2019.

\bibitem{kg_cloud_education_2021}
{IBM Cloud Education}.
\newblock What is a knowledge graph?
\newblock \url{https://www.ibm.com/cloud/learn/knowledge-graph}, Apr 2021.
\newblock [Online; accessed 11-May-2022].

\bibitem{mRAT-SQL}
M.~A. Jose and F.~G. Cozman.
\newblock mrat-sql+gap: A portuguese text-to-sql transformer.
\newblock In {\em Intelligent Systems}, volume 13074 of {\em Lecture Notes in
  Computer Science}, pages 511--525. Springer International Publishing, 2021.

\bibitem{IntegratinQA-NL2SQL}
M.~M. Jos{\'{e}}, M.~A. Jos{\'{e}}, D.~D. Mau{\'{a}}, and F.~G. Cozman.
\newblock {Integrating Question Answering and Text-to-SQL in Portuguese}.
\newblock In {\em International Conference on Computational Processing of the
  Portuguese Language PROPOR 2022: Computational Processing of the Portuguese
  Language}, volume~1, pages 278--287. Springer International Publishing, 2022.

\bibitem{lewis2019bart}
M.~Lewis, Y.~Liu, N.~Goyal, M.~Ghazvininejad, A.~Mohamed, O.~Levy, V.~Stoyanov,
  and L.~Zettlemoyer.
\newblock Bart: Denoising sequence-to-sequence pre-training for natural
  language generation, translation, and comprehension.
\newblock {\em arXiv preprint arXiv:1910.13461}, 2019.

\bibitem{lewis-etal-2021-paq}
P.~Lewis, Y.~Wu, L.~Liu, P.~Minervini, H.~K{\"u}ttler, A.~Piktus, P.~Stenetorp,
  and S.~Riedel.
\newblock {PAQ}: 65 million probably-asked questions and what you can do with
  them.
\newblock {\em Transactions of the Association for Computational Linguistics},
  9:1098--1115, 2021.

\bibitem{li2022data}
B.~Li, Y.~Hou, and W.~Che.
\newblock Data augmentation approaches in natural language processing: A
  survey.
\newblock {\em AI Open}, 2022.

\bibitem{Mallinson2017}
J.~Mallinson, R.~Sennrich, and M.~Lapata.
\newblock Paraphrasing revisited with neural machine translation.
\newblock In M.~Lapata, P.~Blunsom, and A.~Koller, editors, {\em Proceedings of
  the 15th Conference of the European Chapter of the Association for
  Computational Linguistics, {EACL} 2017, Valencia, Spain, April 3-7, 2017,
  Volume 1: Long Papers}, pages 881--893. Association for Computational
  Linguistics, 2017.

\bibitem{miller95}
G.~A. Miller.
\newblock Wordnet: {A} lexical database for english.
\newblock {\em Commun. {ACM}}, 38(11):39--41, 1995.

\bibitem{un2017world}
U.~Nations.
\newblock {\em World Ocean Assessment I}.
\newblock United Nations publication, 2017.

\bibitem{un2021world}
U.~Nations.
\newblock {\em World Ocean Assessment II}.
\newblock United Nations publication, 2021.

\bibitem{plsumeniac}
A.~Oliveira and A.~Costa.
\newblock Plsum: Generating pt-br wikipedia by summarizing multiple websites.
\newblock In {\em Anais do XVIII Encontro Nacional de Inteligência Artificial
  e Computacional}, pages 751--762, Porto Alegre, RS, Brasil, 2021. SBC.

\bibitem{paschoal2021pira}
A.~F. Paschoal, P.~Pirozelli, V.~Freire, K.~V. Delgado, S.~M. Peres, M.~M.
  Jos{\'e}, F.~Nakasato, A.~S. Oliveira, A.~A. Brand{\~a}o, A.~H. Costa, et~al.
\newblock Pir{\'a}: A bilingual portuguese-english dataset for
  question-answering about the ocean.
\newblock In {\em Proceedings of the 30th ACM International Conference on
  Information \& Knowledge Management}, pages 4544--4553, 2021.

\bibitem{pellicer2022ptt5}
L.~F. A.~O. Pellicer, P.~Pirozelli, A.~H.~R. Costa, and A.~Inoue.
\newblock Ptt5-paraphraser: Diversity and meaning fidelity in automatic
  portuguese paraphrasing.
\newblock In {\em International Conference on Computational Processing of the
  Portuguese Language}, pages 299--309. Springer, 2022.

\bibitem{raffel2019exploring}
C.~Raffel, N.~Shazeer, A.~Roberts, K.~Lee, S.~Narang, M.~Matena, Y.~Zhou,
  W.~Li, and P.~J. Liu.
\newblock Exploring the limits of transfer learning with a unified text-to-text
  transformer.
\newblock {\em arXiv preprint arXiv:1910.10683}, 2019.

\bibitem{ramos2003using}
J.~Ramos et~al.
\newblock Using tf-idf to determine word relevance in document queries.
\newblock In {\em Proceedings of the first instructional conference on machine
  learning}, volume 242, pages 29--48. Citeseer, 2003.

\bibitem{reimers2019sentence}
N.~Reimers and I.~Gurevych.
\newblock Sentence-bert: Sentence embeddings using siamese bert-networks.
\newblock {\em arXiv preprint arXiv:1908.10084}, 2019.

\bibitem{Robertson2009}
S.~E. Robertson and H.~Zaragoza.
\newblock The probabilistic relevance framework: {BM25} and beyond.
\newblock {\em Found. Trends Inf. Retr.}, 3(4):333--389, 2009.

\bibitem{Teixeira:2020}
A.~L. Rosa~Teixeira, J.~Campos, R.~Cunha, T.~Castro~Ferreira, A.~Pagano, and
  F.~Cozman.
\newblock {DaMata}: {A} {Robot}-{Journalist} {Covering} the {Brazilian}
  {Amazon} {Deforestation}.
\newblock In {\em Proceedings of the 13th {International} {Conference} on
  {Natural} {Language} {Generation}}, pages 103--106, Dublin, Ireland, Dec.
  2020. Association for Computational Linguistics.

\bibitem{salah2018word}
A.~Salah, M.~Ailem, and M.~Nadif.
\newblock Word co-occurrence regularized non-negative matrix tri-factorization
  for text data co-clustering.
\newblock In {\em Thirty-second AAAI conference on artificial intelligence},
  2018.

\bibitem{scherrer2020tapaco}
Y.~Scherrer et~al.
\newblock Tapaco: A corpus of sentential paraphrases for 73 languages.
\newblock In {\em Proceedings of The 12th Language Resources and Evaluation
  Conference}. European Language Resources Association (ELRA), 2020.

\bibitem{RAT-SQL+GAP}
P.~Shi, P.~Ng, Z.~Wang, H.~Zhu, A.~H. Li, J.~Wang, C.~Nogueira~dos Santos, and
  B.~Xiang.
\newblock Learning contextual representations for semantic parsing with
  generation-augmented pre-training.
\newblock {\em Proc. of the AAAI Conference on Artificial Intelligence},
  35(15):13806--13814, 2021.

\bibitem{souza2020bertimbau}
F.~Souza, R.~Nogueira, and R.~Lotufo.
\newblock Bertimbau: pretrained bert models for brazilian portuguese.
\newblock In {\em Brazilian Conference on Intelligent Systems}, pages 403--417.
  Springer, 2020.

\bibitem{mBART50}
Y.~Tang, C.~Tran, X.~Li, P.~J. Chen, N.~Goyal, V.~Chaudhary, J.~Gu, and A.~Fan.
\newblock {Multilingual Translation with Extensible Multilingual Pretraining
  and Finetuning}.
\newblock {\em arXiv}, 2020.

\bibitem{foreignaffairs2015}
N.~Thompson and R.~Muggah.
\newblock The blue amazon: Brazil asserts its influence across the atlantic.
\newblock {\em Availale at:
  https://www.foreignaffairs.com/articles/africa/2015-06-11/blue-amazon},
  Accessed on 31 May 2022.

\bibitem{UNESCO:2021}
UNESCO.
\newblock Ocean literacy within the {United} {Nations} {Decade} of {Ocean}
  {Science} for {Sustainable} development: a framework for action.
\newblock Technical Report~22, UNESCO Intergovernmental Oceanographic
  Commission, Paris, 2021.

\bibitem{RAT-SQL}
B.~Wang, R.~Shin, X.~Liu, O.~Polozov, and M.~Richardson.
\newblock {RAT-SQL: Relation-aware schema encoding and linking for text-to-SQL
  parsers}.
\newblock {\em arXiv}, 2019.

\bibitem{mT5}
L.~Xue, N.~Constant, A.~Roberts, M.~Kale, R.~Al-Rfou, A.~Siddhant, A.~Barua,
  and C.~Raffel.
\newblock m{T}5: A massively multilingual pre-trained text-to-text transformer.
\newblock In {\em Proc. of the 2021 Conference of the North American Chapter of
  the Association for Computational Linguistics: Human Language Technologies},
  pages 483--498. Association for Computational Linguistics, 2021.

\bibitem{yoo2010orthogonal}
J.~Yoo and S.~Choi.
\newblock Orthogonal nonnegative matrix tri-factorization for co-clustering:
  Multiplicative updates on stiefel manifolds.
\newblock {\em Information processing \& management}, 46(5):559--570, 2010.

\bibitem{autokg}
S.~Yu, T.~He, and J.~R. Glass.
\newblock Constructing a knowledge graph from unstructured documents without
  external alignment.
\newblock {\em CoRR}, abs/2008.08995, 2020.

\end{thebibliography}

\end{document}